\documentclass{bmvc2k}
\usepackage{times}
\usepackage{epsfig}
\usepackage{graphicx}
\usepackage{amsmath}
\usepackage{amssymb}

\usepackage{relsize}
\usepackage{url}
\usepackage{tabularx, booktabs}
\usepackage{pifont}

\usepackage{soul}
\usepackage{multirow}
\usepackage{subcaption}

\definecolor{lightgray}{gray}{0.9}
\definecolor{lightblue}{rgb}{0.93,0.95,1.0}
\definecolor{darkgreen}{rgb}{0.0,0.6,0.0}
\definecolor{mypink1}{rgb}{0.858, 0.188, 0.478}

\newcommand{\minisection}[1]{\vspace{2mm}\noindent{\textbf{#1.}}}
\newcommand{\ignore}[1]{}

\title{Balancing Specialization, Generalization, and Compression for Detection and Tracking}

\addauthor{Dotan Kaufman$^*$}{kaufmand@amazon.com}{1}
\addauthor{Koby Bibas$^*$}{kobibi@amazon.com}{1}
\addauthor{Eran Borenstein}{Eran@amazon.com}{1}
\addauthor{Michael Chertok}{chertokm@amazon.com}{1}
\addauthor{Tal Hassner}{hassner@openu.ac.il}{2}

\addinstitution{
Lab126, Amazon 
}
\addinstitution{
Open University of Israel
}

\runninghead{}{}
\def\eg{\emph{e.g}\bmvaOneDot}

\begin{document}

\maketitle

\begin{abstract}
We propose a method for specializing deep detectors and trackers to restricted settings. 
Our approach is designed with the following goals in mind: (a) Improving accuracy in restricted domains; (b) preventing overfitting to new domains and forgetting of generalized capabilities; (c) aggressive model compression and acceleration. 
To this end, we propose a novel loss that balances compression and acceleration of a deep learning model vs. loss of generalization capabilities. We apply our method to the existing tracker and detector models. 
We report detection results on the VIRAT and CAVIAR data sets. These results show our method to offer unprecedented compression rates along with improved detection. 
We apply our loss for tracker compression at test time, as it processes each video.  
Our tests on the OTB2015 benchmark show that applying compression during test time actually improves tracking performance.
\end{abstract}

\begin{figure}[h!]
\centering
\includegraphics[width=1.0\linewidth]{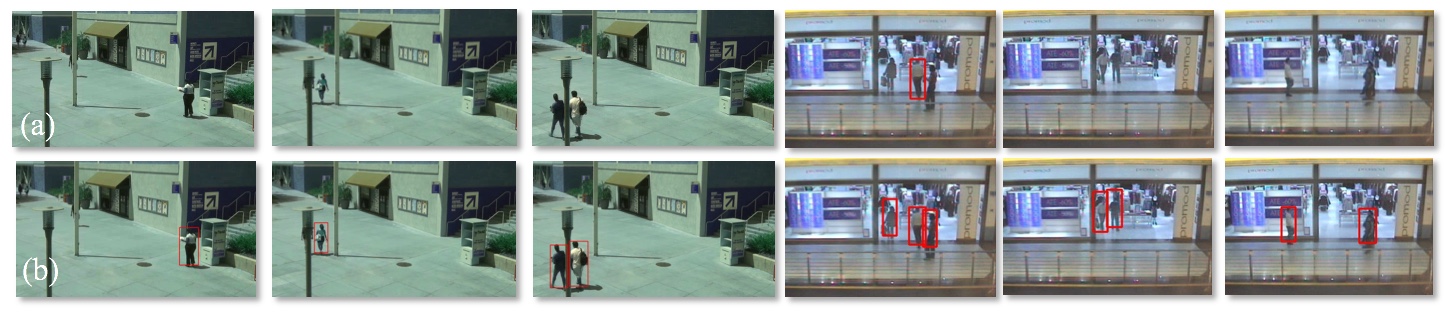}
\vspace{7pt}
\caption{{\bf Qualitative detection results.} 
Result frames from VIRAT videos~\cite{oh2011large} (columns 1-3), and CAVIAR videos~\cite{CAVIAR,fisher2004pets04} (columns 4-6). Row (a) shows detection errors for the general detector. Row (b) shows correct detections for the specialized models for each camera.\vspace{-5mm}}
\label{fig:global_vs_spec}
\end{figure}

\section{Introduction} 
\label{sec:introduction}
Object detection and tracking are fundamental computer vision tasks that have been studied for decades. 
In recent years, most of this research focused on the use of deep neural networks (DNNs). 
Such deep detection and tracking models were since shown to reach remarkable accuracy on a wide range of settings and object appearances.
Though effective, these deep models are also quite expensive in terms of both compute and storage~\cite{hsu2017fastermdnet}. 

In many practical use cases, detection and tracking models are deployed on edge or smart devices. In such settings, the costs associated with running state of the art models can be prohibitive~\cite{howard2017mobilenets,Wang2017FactorizedCN, 2019deep_pnml, goldman2019dense, kaufman2017temporal}. 
We observe that these costs may not be entirely necessary: A model deployed on such devices is often required to process limited input variation. 
An edge-based surveillance camera, for instance, is typically mounted at a fixed location, surveying only one scene throughout its lifespan (Fig.~\ref{fig:global_vs_spec}). 
Moreover, a tracking model follows a single object, tracking it over a sequence of possibly thousands of image frames, all capturing limited variations of backgrounds and object appearances.

We aim to leverage these observations in order to compress and accelerate tracking and detection models in restricted settings, without compromising accuracy. 
A naive approach for compression is to compress a generalized model down to where it may be used on restricted devices~\cite{molchanov2016pruning}. This approach can easily cause overfitting to the target domain if forgetting is not prevented~\cite{serra2018overcoming}. 
Restricting forgetting, on the other hand, can severely limit the potential compression ratios, as we demonstrate in Sec.~\ref{sec:results}.

Alternatively, one can start from a lightweight model and train it on data representing the target domain. 
Light enough models can then be used on the restricted device. One example of such an approach is the recent MobileNets~\cite{howard2017mobilenets}. 
Comparing the performance of a MobileNets detector with its heavy~\cite{SSD} counterpart shows that accuracy is lost~\cite{howard2017mobilenets}.

We present a method for {\em compressing and accelerating} models trained for general-purpose tracking and detection, by specializing the models to restricted environments. We offer the following technical contributions. 
\begin{enumerate}
\item {\bf Model specialization vs. generalization.} We propose a loss that optimizes the balance between specialization, generalization, and compression of our detection and tracking models.
\item {\bf Model compression vs. acceleration.} Different networks of the same disk size can have very different computational requirements. We, therefore, design a loss that encourages the creation of architectures that minimize computation time.
\item {\bf Run time tracker compression and acceleration.} We compress and accelerate a tracker {\em at test time}, obtaining state of the art performance with a model roughly half the size of the original. 
\end{enumerate}

We apply our method to a pedestrian detector~\cite{SSD} trained on the PASCAL~\cite{pascal-voc-2012} dataset, representing generalized detection domains. 
Each of the security cameras in VIRAT~\cite{oh2011large} and CAVIAR~\cite{CAVIAR} represent a separate restricted domain. 
For tracking, we show results on the OTB2015~\cite{wu2015object} dataset using MDNet~\cite{nam2016learning}.

Our results in Sec.~\ref{sec:results} show that a general detector can be accelerated by $\times 5.6$ while maintaining the same performance on the restricted domain. 
Moreover, our novel loss enables the use of self-supervision in the restricted domain, which improves the general detector by up to $\times$1.3 in average precision. 
Additionally, we show that a tracker model can be accelerated by $\times$3, with a drop of only 2\% in performance. 
On lower compression rates we achieve a performance {\em boost} of 2\% over the initial state of the art tracker.


\section{Related work} \label{sec:related}
\minisection{Object detection} 
Single-image object detectors are often grouped into two categories. {\em Region-based / two-stage methods} (\eg, R-CNN~\cite{girshick2014rich} and others~\cite{he2017mask, ren2015faster}) detect objects by proposing object regions and then classifying each region. {\em Single-stage approaches}, on the other hand, generate predictions at fixed {\em anchor} positions~\cite{dai2016r,SSD}.
Another approach, called Mask R-CNN, extends Faster R-CNN \cite{he2017mask} by adding a branch for predicting segmentation masks
on each Region of Interest (RoI), in parallel with the existing branch for classification and bounding box regression.  

\minisection{Object tracking} 
Developing visual trackers typically involves the design of representations, localizing methods, and online appearance update schemes. 
Appearance representations are often captured by deep neural networks that were pretrained for recognition tasks. 
Some examples of this approach are MDNet~\cite{nam2016learning} and CREST~\cite{song2017crest}.

\minisection{Pruning in the context of domain adaptation}
Deep networks are often pruned by setting weights that are lower than some threshold to zero, thereby producing sparse weight matrices which are easy to compress~\cite{han2015learning}.
Other methods instead cut neural connections to produce small, dense networks~\cite{he2017channel,molchanov2016pruning}. 
Our design is agnostic to the particular pruning method used. 
We focus instead on the application of pruning as part of the model specialization. 
Importantly, although self-supervised domain adaptation and model compression were studied in the past, they were rarely tackled together, as we do here.

Serra et al.~\cite{serra2018overcoming} proposed a new compression method that prunes convolutional filters.
Our method is inspired by their work. However, their design does not aim for runtime acceleration. 
Our method, on the other hand, takes into account the compute costs directly, thus, it can be used when fast inference time is the desired property of the specialized model.

\section{Specializing with controlled forgetting} \label{sec:novelty}
Our goals for specializing the DNN models to a restricted domain are as follows: 
\begin{enumerate}
    \item{\bf Specialization.} Improve accuracy in a restricted, target domain.
    \item{\bf Generalization.} Prevent catastrophic forgetting of information learned on a generalized training set and thereby avoid overfitting to the target domain.
    \item{\bf Compression and Acceleration.} Compress the network disk footprint and reduce run time compute.
\end{enumerate}
We express these goals formally as loss functions used to specialize the model. 
These losses weigh, in an adaptive manner, the influences of training data representing the original (generalized) settings vs. the restricted domain.

\subsection{Preliminaries: detector and tracker}
\label{sec:preliminary}
We first describe the (existing) detection and tracking methods used. Importantly, we chose both components due to convenience: We can apply the same specialization method with other choices of detectors and trackers. As we later report in Sec.~\ref{sec:results}, even though the detector used here is no longer necessarily state of the art~\cite{wang2017fast}, our optimization improves its accuracy to the point where it {\em outperforms the best results reported by others}.

\minisection{Tracker}
MDNet~\cite{nam2016learning} is a popular DNN-based tracking algorithm with state-of-the-art accuracy. To track a given object, it samples candidate regions around the object, which are passed through a DNN pretrained on a large-scale dataset and fine-tuned at the first frame in a test video. Target locations in the subsequent frames are determined by averaging the bounding box outputs of top scoring positive patches. It collects positive and negative samples during the tracking process, and regularly updates the classifier.

\minisection{Detector}
In our setup, we use the SSD detector~\cite{SSD}, as we found it to offer a good trade-off between accuracy and speed. 
Briefly, SSD uses a base feature extraction network, VGG-16~\cite{simonyan2014very}, followed by convolutional layers that predict (refined) locations $\mathbf{d}$ and objectness confidences, $\mathbf{c}(\mathbf{d})$, for a large set of regions, predefined to cover the image domain in multiple scales and locations. The initial, generalized SSD is trained offline for generalized pedestrian detection on PASCAL data set (Sec~\ref{sec:data}). During training, we use the original loss and optimization procedure defined for SSD~\cite{SSD}. 

\minisection{Tracking as a supervisory detection signal} We consider settings where a camera captures videos in a restricted environment. Our approach uses an existing detector, trained on a generalized detection data set and an object tracker. Both detector and tracker localize bounding boxes for objects in video frame $\mathbf{I}^t$. Bounding boxes are defined as 4-tuples, $\mathbf{b}=(x,y,h,w)$: the 2D coordinate of the center of the box, its height, and width respectively. 

Detection and tracking methods provide confidence scores for each bounding box, $\mathbf{c}(\mathbf{b})\in[0, 1]^p$. These confidences represent the classification probability of the box to each of $p$ object categories. Thresholds are applied to these scores and only bounding boxes with at least one confidence higher than the threshold are returned. In the restricted settings, we detect only a single class (pedestrians). We base our tracker on MDNet. We use the SSD detector to initialize new {\em tracklets}, representing tracked object trajectories. 
The tracker searches for a bounding box in $\mathbf{I}^{t+1}$ that best matches a bounding box in $\mathbf{I}^t$.

To manage multiple, possibly concurrent tracklets, we implemented an {\em object manager unit}, similar to one proposed by others~\cite{chu2017online}. This object manager combines the output of our detector and tracker to determine birth, refinement, and termination of all tracklets.

Tracklets collected by the object manager unit are used as a source of training data for specializing the detector to its new domain~\cite{andriluka2008people,mao2015training} which is in our experiments represented by clips taken from security cameras datasets: VIRAT \cite{oh2011large}, CAVIAR \cite{CAVIAR,fisher2004pets04}. Importantly, we use only the data from the tracker to train our restricted detector model on these sets.

\subsection{Specialization Vs. Generalization} \label{sec:novelty:data}
In order to balance specialization vs. generalization, we introduce two sets: one representing the generalized domain and the other representing the restricted domain. 
The training data from the general domain, $\mathbf{D}^{G}$, contains images and their supervised ground truth annotations. 
To adapt a generalized detection and tracking models we use a self-supervised approach of collecting training samples with a tracker. 
We refer to this set of samples which will be used to train a specialized model on a restricted domain as $\mathbf{D}^\textit{R}$. 
We use the model with its original loss function to define the specialization loss, the loss on the restricted domain: $\mathcal{L}_{S} = \mathcal{L}(\mathbf{D}^{\textit{R}})$.

While specializing the DNN model, we wish to retain as much of its generalized capabilities as possible. 
Unlike others~\cite{serra2018overcoming}, we allow the network to forget some of its previous knowledge, favoring a soft remembrance mechanism. We do this by training the specialized model with samples taken from the supervised set. 
That is, where $\mathbf{D}^{G}$ is a set taken from the generalized training set, we use $\mathcal{L}_{G}  = \mathcal{L}(\mathbf{D}^{G})$ as the generalized loss.

Preserving model performance on the generalized set is important due to two reasons. 
Firstly, due to the nature of the restricted domain data set, which is collected in a self-supervised manner
and may contain noisy labels which could reduce the performance on the restricted domain. 
Secondly, maintaining performance on the generalized set serves as a measure to avoid overfitting to the restricted domain.

\subsection{Compression and acceleration} \label{sec:novelty:compression}
Our approach can utilize different compression methods. Unlike others, we wish to accelerate network runtime, not only compress its size. Thus, we favor methods which remove entire filters, rather than individual weights.
Our goal is to reduce the inference time, hence, we would like to remove the filters taking the most of the floating point operations (FLOPS).

We define {\em embedding units}, $\{e_{l,i}\}_{i=1}^{N_l}$, for each of the $\{N_l\}_{l=1}^L$ convolution filters in a network which consists of $L$ layers. When forwarding a sample in the model, each convolution filter is multiplied by its corresponding mask, $\sigma(e_{l,i})$, where $\sigma(\cdot)$ is a Sigmoid gate. 

We note that the number of FLOPS computed in the l-th layer of a convolutional DNN is given  by $FLOP_l=N_{l-1} N_l F_l K_l$,
where $F_l$ and $K_l$ are the feature map size and the convolution kernel size respectively.
After applying Sigmoid gates the effective number of convolution kernels becomes $\hat{N_l}=\sum_{i=1}^{N_l}  \sigma(e_{l,i})$ 
which indicates that the effective FLOPS number is
$\hat{FLOP}_l= \hat{N}_{l-1} \hat{N}_l F_l K_l$.

We suggest a new loss, specifically designed to optimize compute costs by reducing the number of FLOPS
\begin{equation} \label{eq:loss:compression_gflops}
\mathcal{L}_C = \frac{1}{FLOPS} \sum_{l=1}^L \hat{\textit{FLOP}}_l.
\end{equation}
where $FLOPS=\sum_{l=1}^{L}\textit{FLOPS}_l$ is the model's total FLOPS and is used for normalization.

Minimizing Eq.~\eqref{eq:loss:compression_gflops} is equivalent to minimizing the effective number of FLOPS~\cite{friedman2001elements} and therefore producing a sparse mask over the convolution filters.
Filter with a mask value close to zero creates a low-value activation and can, therefore, be removed without influencing the overall performance of the network. By removing the filter, we reduce the model size as well as accelerate runtime.

\subsection{The combined loss}\label{sec:combinedloss}
Using the losses from Sec.~\ref{sec:novelty:data}--Sec.~\ref{sec:novelty:compression} we define the combined objective function used to specialize the DNN:
\begin{equation} \label{eq:loss:final}
    \mathcal{L}(\beta,\lambda) = \mathcal{L}_S + \beta \mathcal{L}_{G} + \lambda \mathcal{L}_{C}. 
\end{equation}
The coefficient $\beta$ controls how much of the previous, generalized task the specialized model will remember. 
The values of $\lambda$ determine the compression of the specialized model. 
Our experiments provide an in-depth analysis of the values of $\beta$ and $\lambda$ and their effects.

\section{Experiments} \label{sec:results}
\subsection{Data sets} \label{sec:data}
Our tests focus on settings where specialization can be assumed. To that end, we conduct our tests on data sets representing restricted domains. Our detection tests use footage taken from surveillance cameras, provided by standard surveillance data sets, VIRAT Ver 2.0~\cite{oh2011large} and CAVIAR Ver 2.0~\cite{CAVIAR}. Our tracking tests specialize the same tracker, each time to a different video; each video representing a separate restricted domain. We use the public OTB2015 tracking benchmark for this purpose~\cite{wu2015object}.

\minisection{VIRAT Ver 2.0~\cite{oh2011large}} This set contains data acquired from stationary video cameras, typically located on building rooftops. The set includes approximately 25 hours of video footage taken at 16 different scenes. In our experiments, we used the first 60\% of the unlabeled videos from each camera as our training set. The remaining 40\% was used for testing along with their ground truth supervised bounding box annotations.

\minisection{CAVIAR Ver 2.0~\cite{CAVIAR}} This set contains 52 annotated video sequences with a total of 90K frames, of which 52K target frames provide ground truth annotations. About one-third of these videos capture an indoor office lobby. The rest of the videos were taken at an indoor shopping center. We used 60\% of the videos for training, evenly selected across all scenes. The remaining videos, along with their ground truth annotations, were used for testing.

\minisection{PASCAL~\cite{pascal-voc-2012}} The PASCAL set contains around 10K images with their labeled bounding boxes. From this set, we randomly sampled 3800 training images that contain instances of the {\em person} class. We additionally used 500 test images from the test set to measure the performance of the specialized detector on the generalized domain.

\minisection{OTB2015~\cite{wu2015object}} This a popular object tracking data set which consists of 100 fully annotated videos with various challenging attributes. It contains sequences where the object vary in scale and has different illumination. 

\subsection{Evaluation methodology}\label{sec:evalmethod}
We report detection performance using the following evaluation metrics: 

\minisection{\#GFLOPS Ratio} The ratio between the number of the original general model GFLOPS $\#\textit{GFLOPS}_1$ and the specialized model GFLOPS $\#\textit{GFLOPS}_2$, i.e., $\frac{\#\textit{GFLOPS}_1}{\#\textit{GFLOPS}_2}$.

\minisection{Improvement in $AP^{R}$ ($\times AP^{R}$)}
The ratio between the specialized model mean average precision to the generalized model mean average precision at $IoU=0.5$. This ratio is measured on the restricted test set, comparing the specialized with the generalized model. Values larger than $1.0$ imply improvement while values below $1.0$ imply degradation.  

\minisection{Improvement in $AP^{G}$  ($\times AP^{G}$)}
The ratio between the specialized model mean average precision to the generalized model mean average precision at $IoU=0.5$. This ratio is measured on the generalized test set, comparing the specialized with the generalized model. 

\minisection{\%Compression} The ratio between the number of the pruned parameters of the specialized model to the number of parameters of the generalized model.

\vskip 0.2cm
\noindent We report tracking performance using the \%Compression rate, \#GFLOPS Ratio, and the following evaluation metrics:

\minisection{Precision} Precision score is measured as the percentage of frames
whose predicted object location (center of the predicted
box) is within a distance of 20 pixels from the center of
the ground truth box. 

\minisection{Success} To measure the performance on a sequence of frames, we count the number of
successful frames whose $IoU$ is larger than a threshold. 
The success plot shows the ratios of successful frames as the thresholds varies from 0 to 1. We use the area under the curve (AUC) of each success plot to rank a tracker.

\subsection{Baseline compression methods} \label{sec:baselines}
We test our method using various compression techniques:

\minisection{Zeroing~\cite{han2015learning}} A compression scheme where model weights which are lower than some threshold are set to zero. In this case, when computing the \#GFLOPS Ratio, only convolution kernel parameters greater than zero are taken into account.

\minisection{Taylor~\cite{molchanov2016pruning}} 
In this method, we feed-forward the training set and then perform back-propagation. For each filter, a rank is created by multiplying the filter's matching activations with their gradients.
Filters with low rank are then pruned to create a smaller model.

\minisection{Hard Attention to the Task (HAT)~\cite{serra2018overcoming}} This compression scheme involves a similar filter embedding method, as we proposed in Sec.~\ref{sec:novelty:compression}. Their approach, however, does not optimize for model runtime.

\minisection{FLOPS Loss} Our compression method of Sec.~\ref{sec:novelty:compression}, which is designed to reduce model size and compute cost.

\begin{figure}%
    \centering
    \begin{subfigure}[t]{0.24\linewidth}
        \includegraphics[width=1\linewidth]{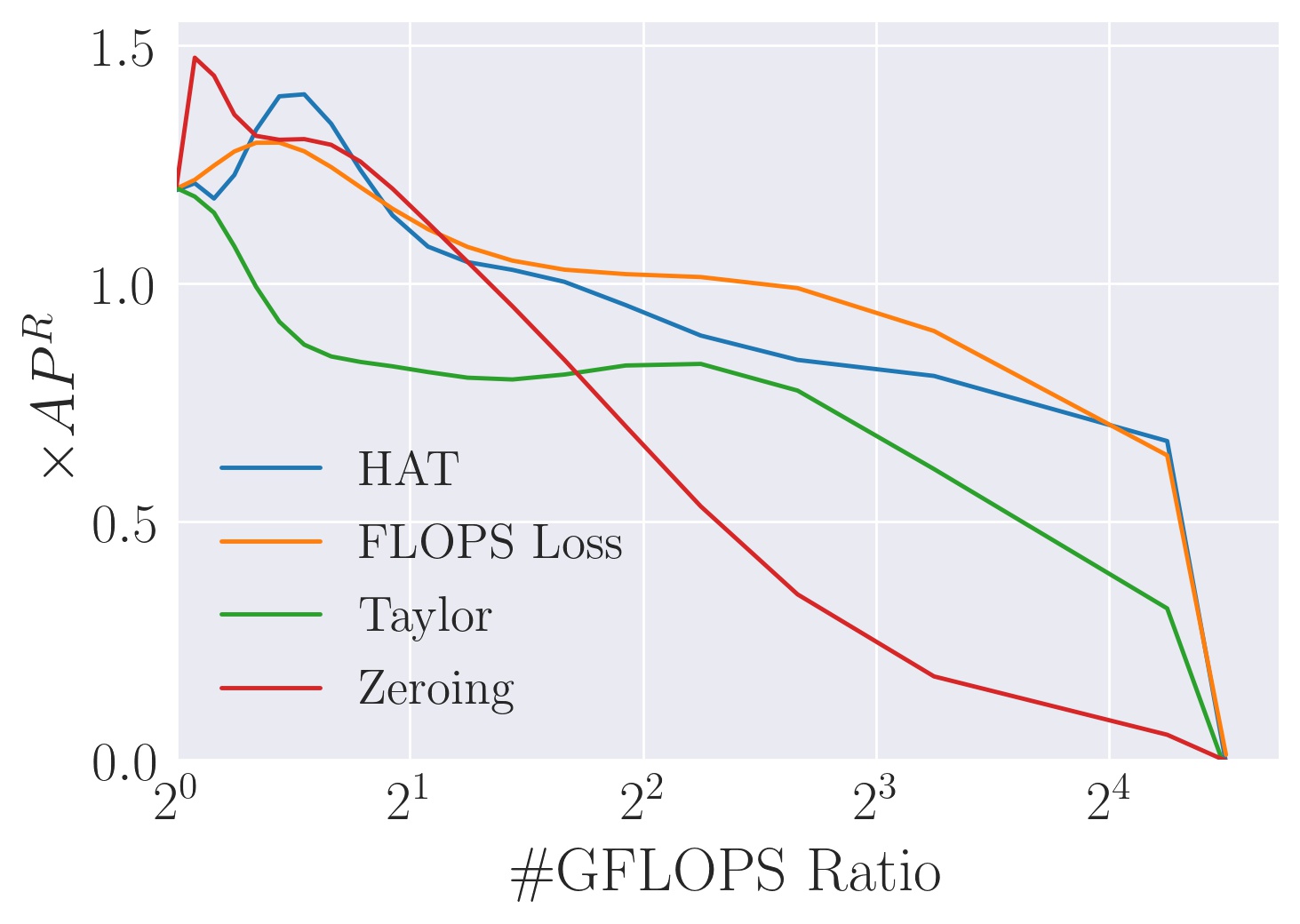}
        \caption{VIRAT $\times AP^R$. \label{fig:virat_restricted}}
    \end{subfigure}
    \begin{subfigure}[t]{0.24\linewidth}
        \includegraphics[width=1\linewidth]{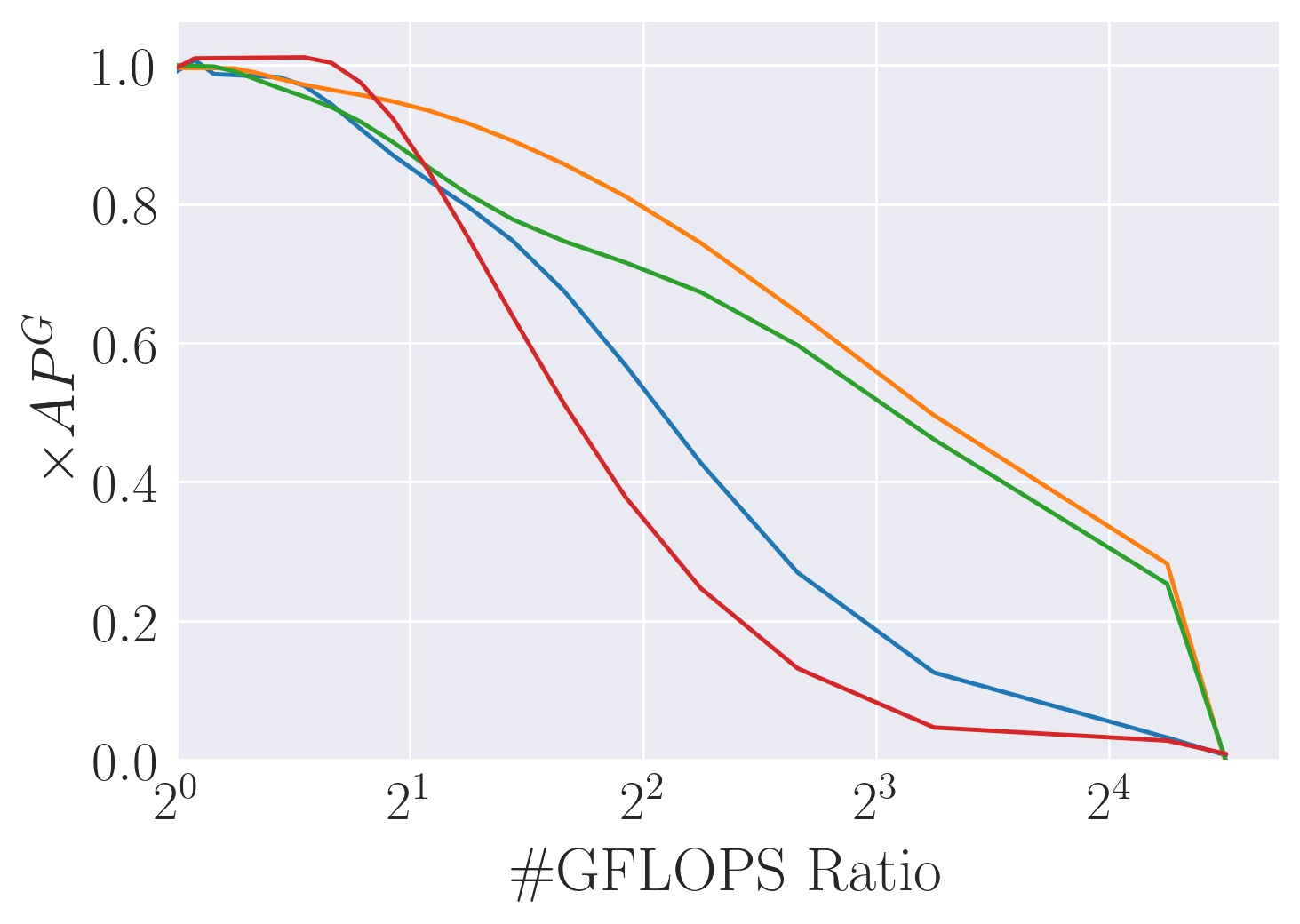}
        \caption{VIRAT $\times AP^G$. \label{fig:virat_generalized}}
    \end{subfigure}
    \begin{subfigure}[t]{0.24\linewidth}
        \includegraphics[width=1\linewidth]{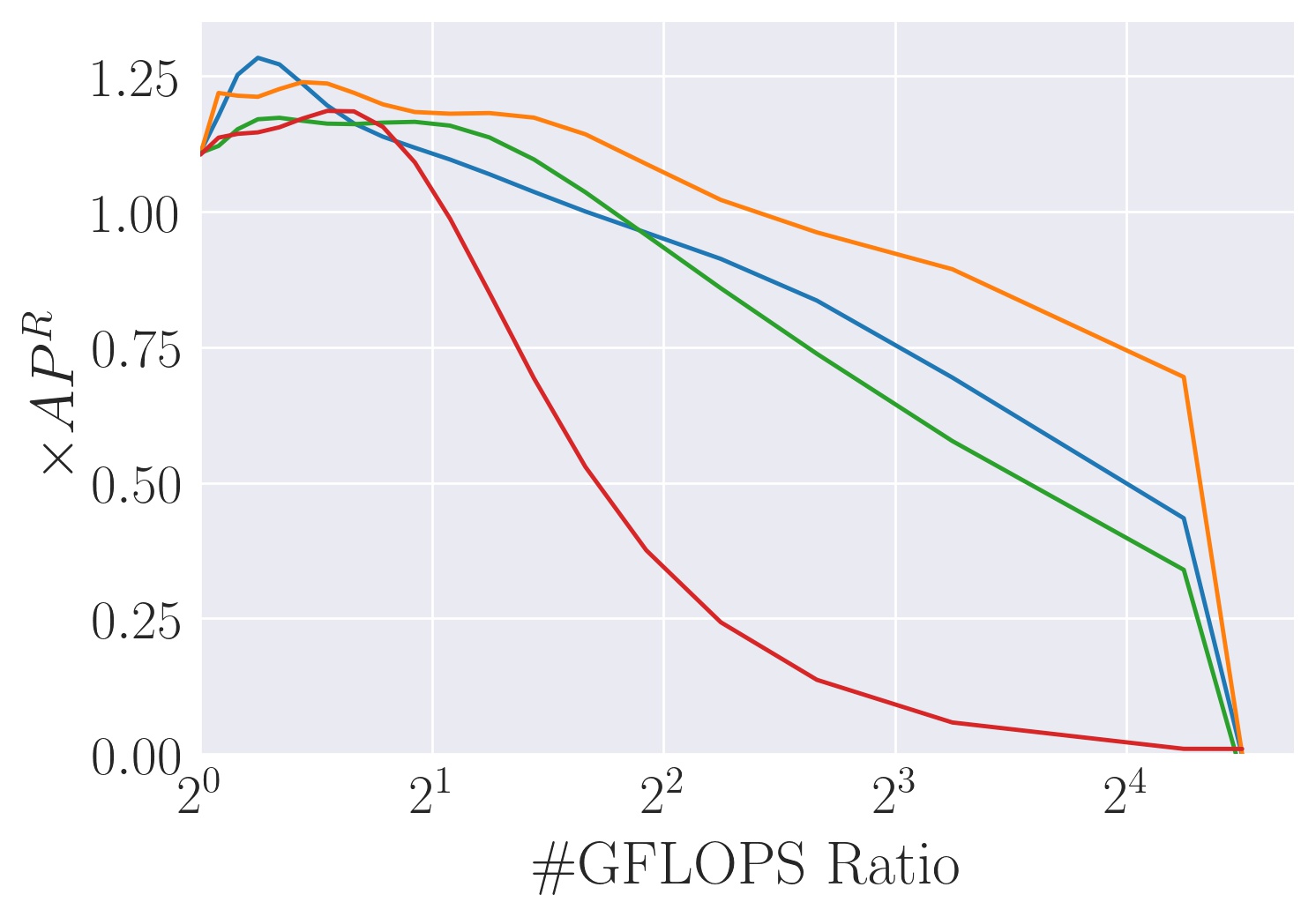}
        \caption{CAVIAR $\times AP^R$. \label{fig:caviar_restricted}}
    \end{subfigure}
    \begin{subfigure}[t]{0.24\linewidth}
        \includegraphics[width=1\linewidth]{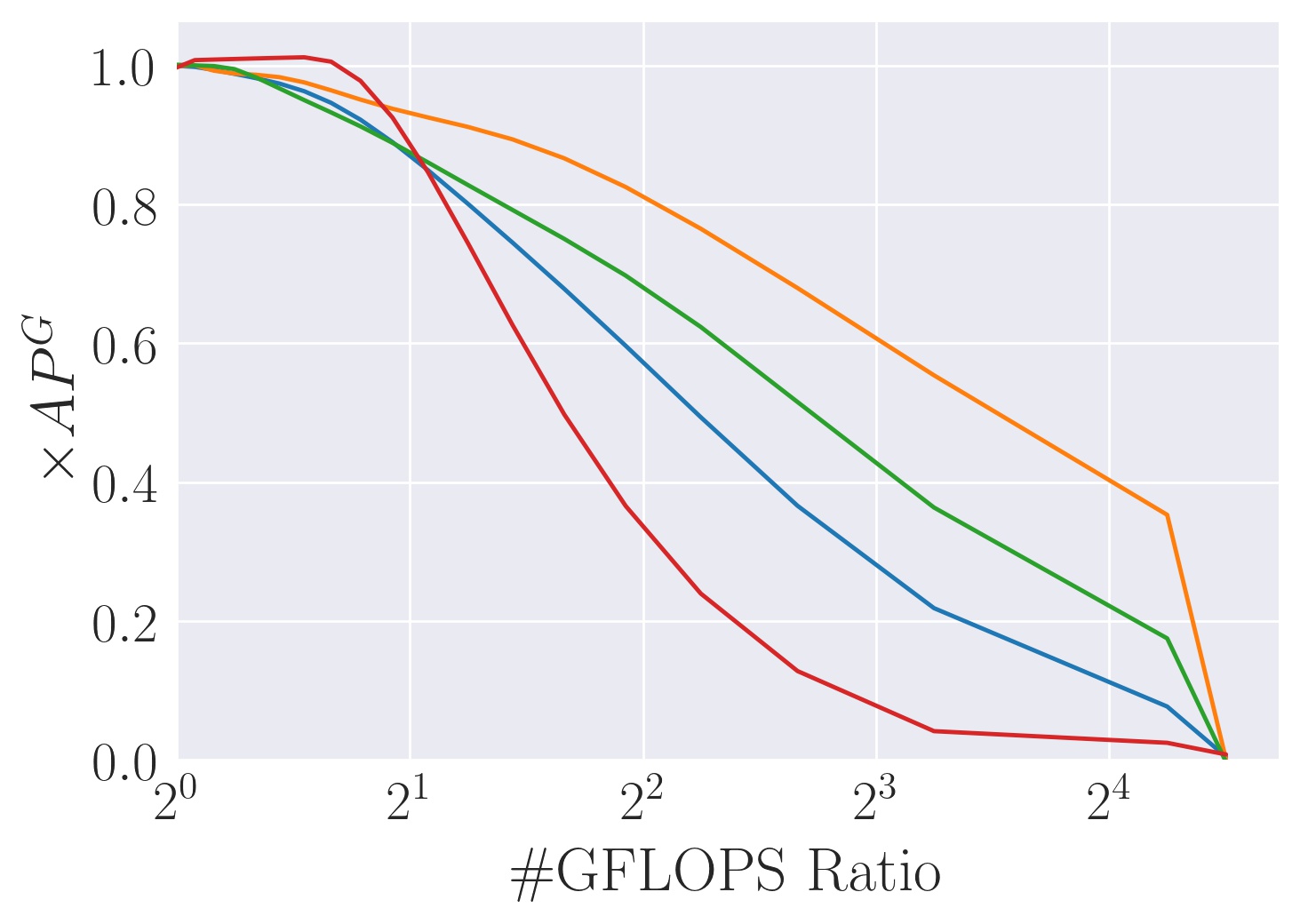}
        \caption{CAVIAR $\times AP^G$. \label{fig:caviar_generalized}}
    \end{subfigure}
    \vspace{10pt}
    \caption{\textbf{Specialized model performance.} Pruning results on VIRAT and CAVIAR test sets. We tested several compression methods, see Sec. (\ref{sec:baselines})}.
    \vspace{-5mm}
\end{figure}

\begin{table}[tb]
\centering
\resizebox{0.75\linewidth}{!}{%
\begin{tabular}{lccccc} 
\toprule
Method                                 & $\times AP^R$     &  $\times AP^G$    & \%Compression   &	$\times AP^R$ AUC           &  $\times AP^G$ AUC 	    \\
\bottomrule
\multicolumn{6}{c}{{\bf VIRAT}~\cite{oh2011large}}      \\
Zeroing~\cite{han2015learning}          &   0.43           &   0.18            &   9.60         &   6.23                        &   9.53                    \\
Taylor~\cite{molchanov2016pruning}      &   0.80           &   0.63            &   \textbf{34.72}        &   11.59                       &   9.53 	                \\
HAT~\cite{serra2018overcoming}          &   0.86           &   0.34            &   9.14         &   16.17                       &   4.70                    \\
FLOPS Loss \eqref{eq:loss:compression_gflops}  &  \textbf{1.00}     &   \textbf{0.69}            &   25.37        & \textbf{17.18}                & \textbf{10.39}            \\
\bottomrule										
\multicolumn{6}{c}{{\bf CAVIAR Ver 2.0}~\cite{CAVIAR,fisher2004pets04}} \\
Zeroing~\cite{han2015learning}          & 0.18              &   0.18            & 9.92          & 3.59                          &  3.27	                    \\
Taylor~\cite{molchanov2016pruning}      & 0.79              &   0.56            & \textbf{33.90}         & 12.19                         &  8.14                     \\
HAT~\cite{serra2018overcoming}          & 0.87              &   0.42            & 7.50          & 13.87                         &  5.92                     \\
FLOPS Loss \eqref{eq:loss:compression_gflops}  & \textbf{0.99}       &   \textbf{0.72}            & 26.57	        & \textbf{17.6}                 & \textbf{11.35}            \\
\bottomrule
\end{tabular}%
}
\vspace{3mm}
\caption{\textbf{Specialized detector performance.} 
The three left columns show the performance of the models that were specialized on the CAVIAR and VIRAT sets at \#GFLOPS Ratio of $5.6$.
The two right columns are the matching AUC values of Fig.~\ref{fig:virat_restricted},  \ref{fig:virat_generalized}, \ref{fig:caviar_restricted}, \ref{fig:caviar_generalized} for the various compression methods that were used for specialization.}
\label{table:restricted_ap_auc} \label{table:specialized_detecor_performance}
\vspace{-3.5mm}
\end{table}

\subsection{Compression analysis} 

\noindent{\bf Detection tests.} We tested our specialization method on the VIRAT and CAVIAR benchmarks. We start with a general object detector which was trained on the PASCAL data set. 
We then perform fine-tuning using the self-supervised data which was collected by the tracker. 
After which, we apply our method using Eq.~\eqref{eq:loss:final}. We begin by analyzing the influence of $\lambda$ while setting $\beta=1$. We test $\lambda$ values in the range of $0-100$. 

Qualitative results of the specialized detector model on the VIRAT and CAVIAR sets are shown in Fig.~\ref{fig:global_vs_spec}. Fig.~\ref{fig:virat_restricted} presents the results of our method using several compression techniques on the VIRAT set. As can be seen, fine-tuning improves detection performance on the restricted domain. Our method improves performance further, while successfully reducing compute costs. Remarkably, using our framework we are able to maintain the same performance as the initial, general detector at a \#GFLOPS ratio of 5.6.   
Fig.~\ref{fig:virat_generalized} presents the performance of the specialized model on the PASCAL data set. We see a monotonic decrease in performance which indicates that the more compressed the model, the more it tends to forget.

Fig.~\ref{fig:caviar_restricted} shows the performance of the specialized model on the CAVIAR set. 
Our method is able to maintain the same performance as the initial, general detector at a \#GFLOPS ratio of 5.6. Fig.~\ref{fig:caviar_generalized} shows the performance of the specialized model on the PASCAL data set. Our FLOPS Loss method achieves the best performance on this set. 

In order to compare our proposed FLOPS Loss method with other compression methods, for different values of GFLOPS, we use the AUC (area under the curve).
Quantitative results of $\times AP^R$ and $\times AP^G$ AUCs for both data sets are presented in Table.~\ref{table:restricted_ap_auc}.
Our FLOPS Loss compression method (Eq.~\eqref{eq:loss:compression_gflops}) outperforms all other compression methods.
Moreover, our suggested loss has the best performance on the generalized set with $\times AP^G$ AUC of 10.39 on the VIRAT set and $\times AP^G$ AUC of 11.35 on the CAVIAR set.
Table.~\ref{table:specialized_detecor_performance} demonstrates the performance of the specialized detectors at \#GFLOPS ratio of $5.6$. At this point, the specialized detector performance is as good as the original detector ($\times AP^R$ of 1.0). Our suggested compression FLOPS Loss has the best $\times AP^R$ and $\times AP^G$ performance at this point. HAT compression method attains better \%Compression due to the nature of its compression loss which prefers smaller size models and not reducing compute costs like our method.

\begin{figure}
    \centering
    \begin{subfigure}[t]{0.32\linewidth}
        \includegraphics[width=1\linewidth]{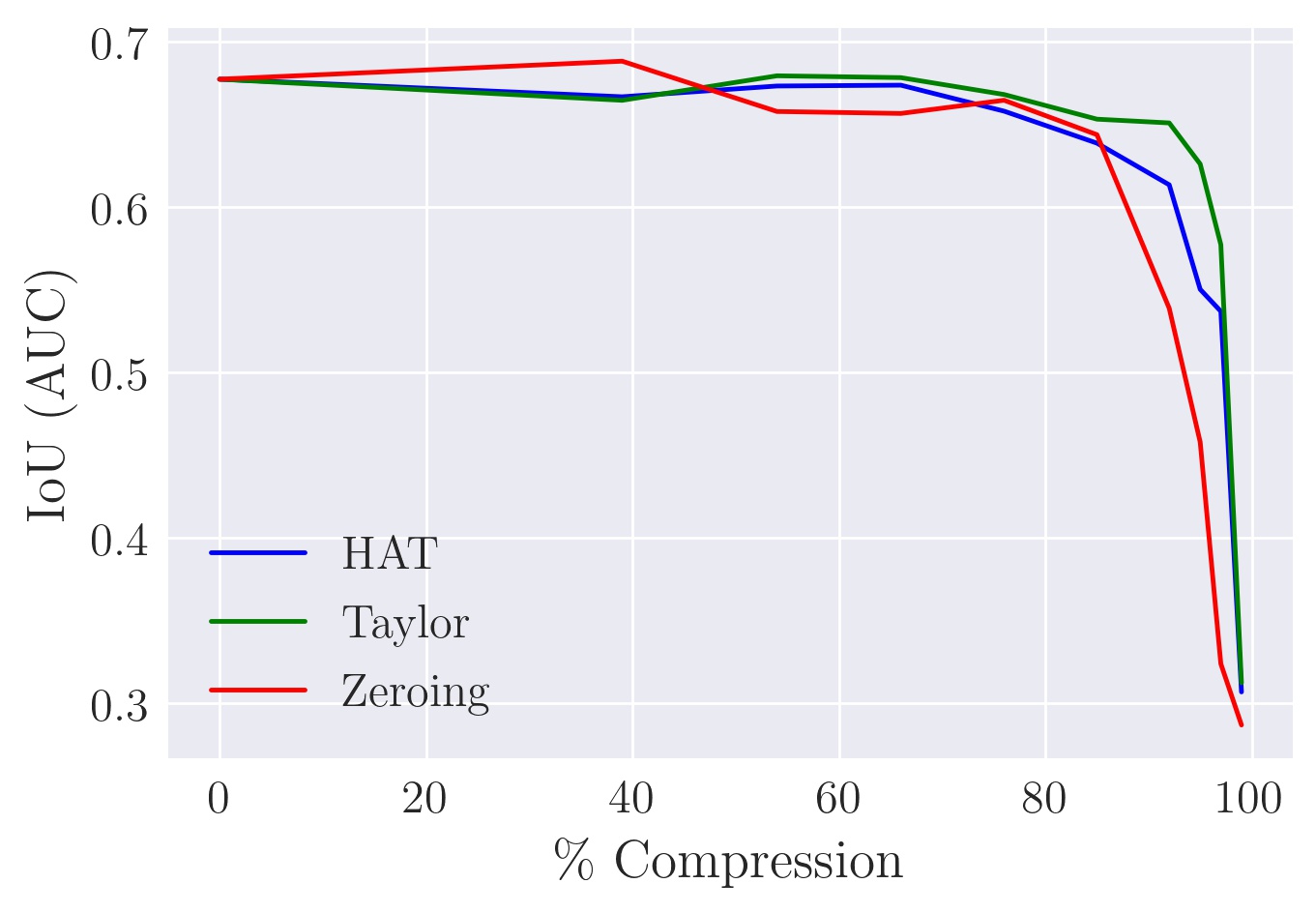}
        \caption{Success AUC. \label{fig:otb_precisicn_omopression}}
    \end{subfigure}
    \begin{subfigure}[t]{0.32\linewidth}
        \includegraphics[width=1\linewidth]{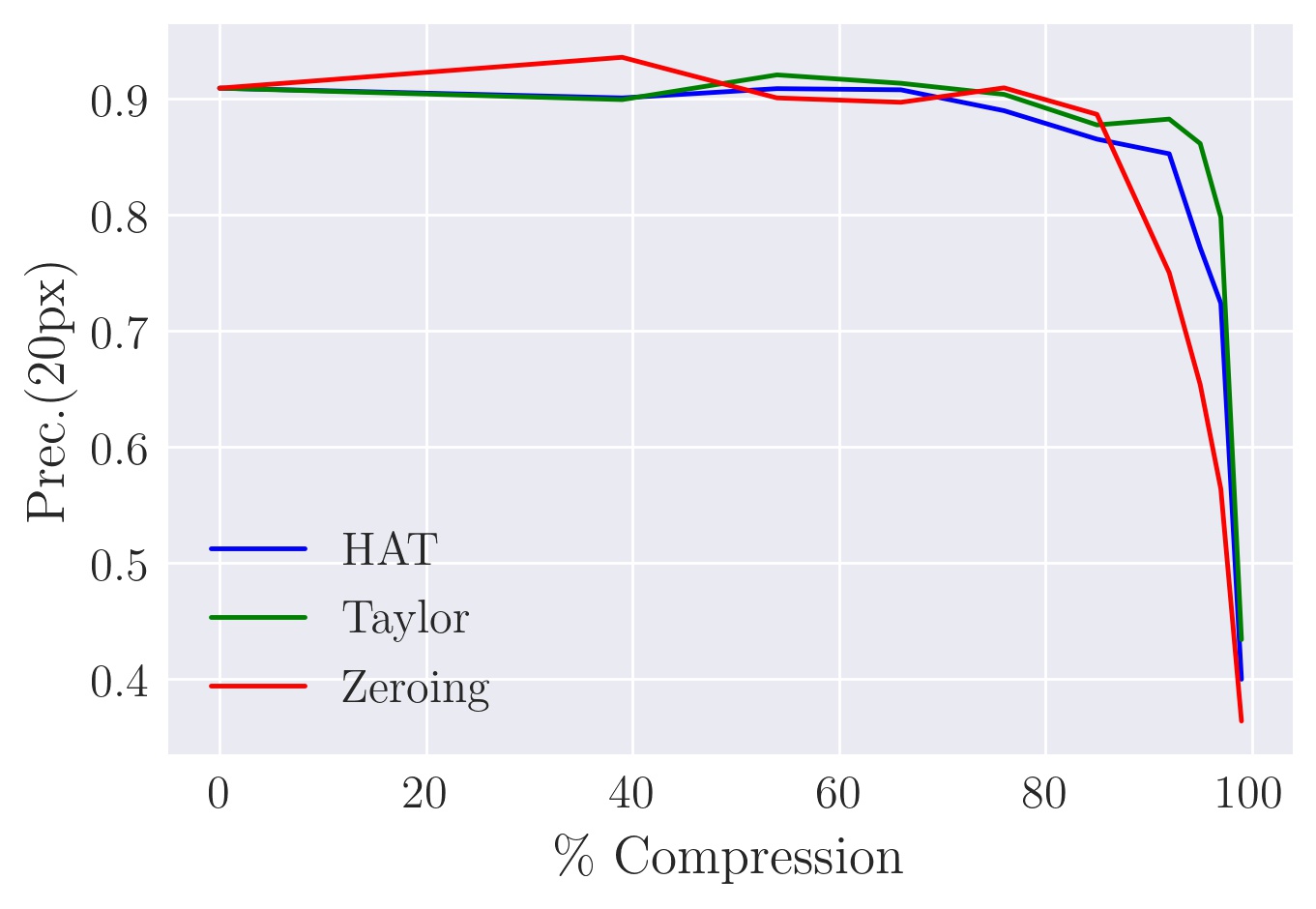}
        \caption{Precision@20 score. \label{fig:otb_success_compression}}
    \end{subfigure}
    \begin{subfigure}[t]{0.32\linewidth}
        \includegraphics[width=1\linewidth]{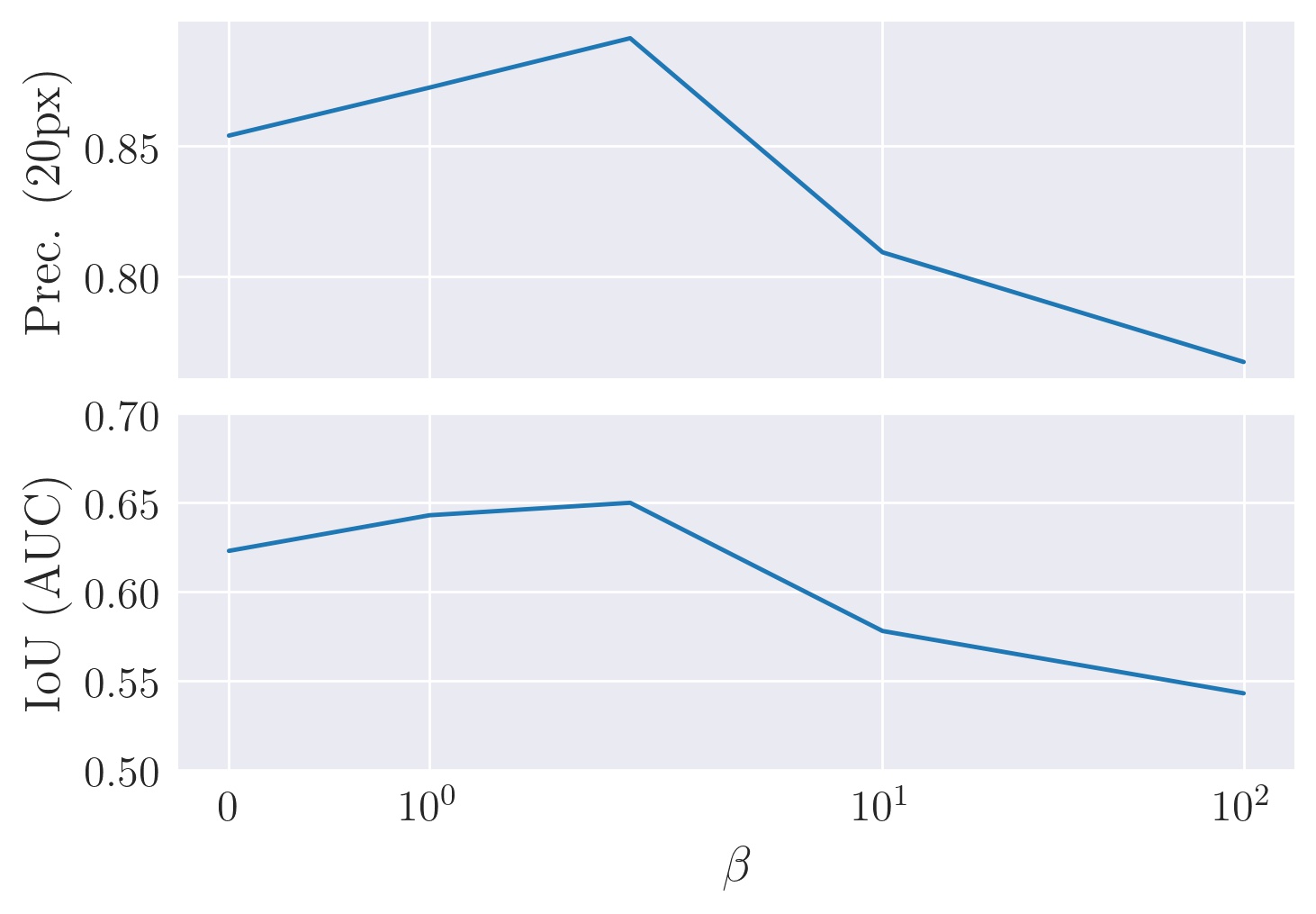}
        \caption{$\beta$ evaluation. \label{fig:beta_tracker}}
    \end{subfigure}
    \vspace{10pt}
    \caption{\textbf{OTB2015 Results} The left and middle figures show the performance of the specialized tracker for different pruning methods (Sec.~\ref{sec:baselines}) on the OTB2015 set.
    The right figure shows the success AUC and precision scores for a compression rate of $\times$13,
    for several values of $\beta$, while using the Taylor compression method.}
\end{figure}

\minisection{Tracking tests}
At inference, MDNet tracker manner of operation consists of training with positive and negative samples drawn from the initial frame of the clip. Additionally, as mentioned in Sec. \ref{sec:preliminary}, MDNet regularly trains using data it collects in subsequent frames. 
Our method prunes MDNet using
Eq. \eqref{eq:loss:final}, at test time.
Importantly, we found that this does not require a change in the number of training iterations, thus, not adding any computational overhead. As can be seen from  Fig.~\ref{fig:otb_precisicn_omopression},~\ref{fig:otb_success_compression}, we found Taylor pruning method to perform better at high compression rates. We set $\beta=2$ in our experiments. 
Using Zeroing we are able to zero out 40\% of the network while improving its precision score by 3\%, and its success score by 2\%. 

When evaluating a pretrained compressed model where pruning was performed at the pre-training stage, we observe a significant drop in performance. Specifically, the precision score drops by 15\% while the success drops by 21\% at a compression rate of $\times$13 (compression of 92\%). This validates the importance of compression in the restricted domain.

\begin{figure}
    \centering
    \begin{subfigure}[t]{0.49\linewidth}
        \includegraphics[width=1.0\linewidth]{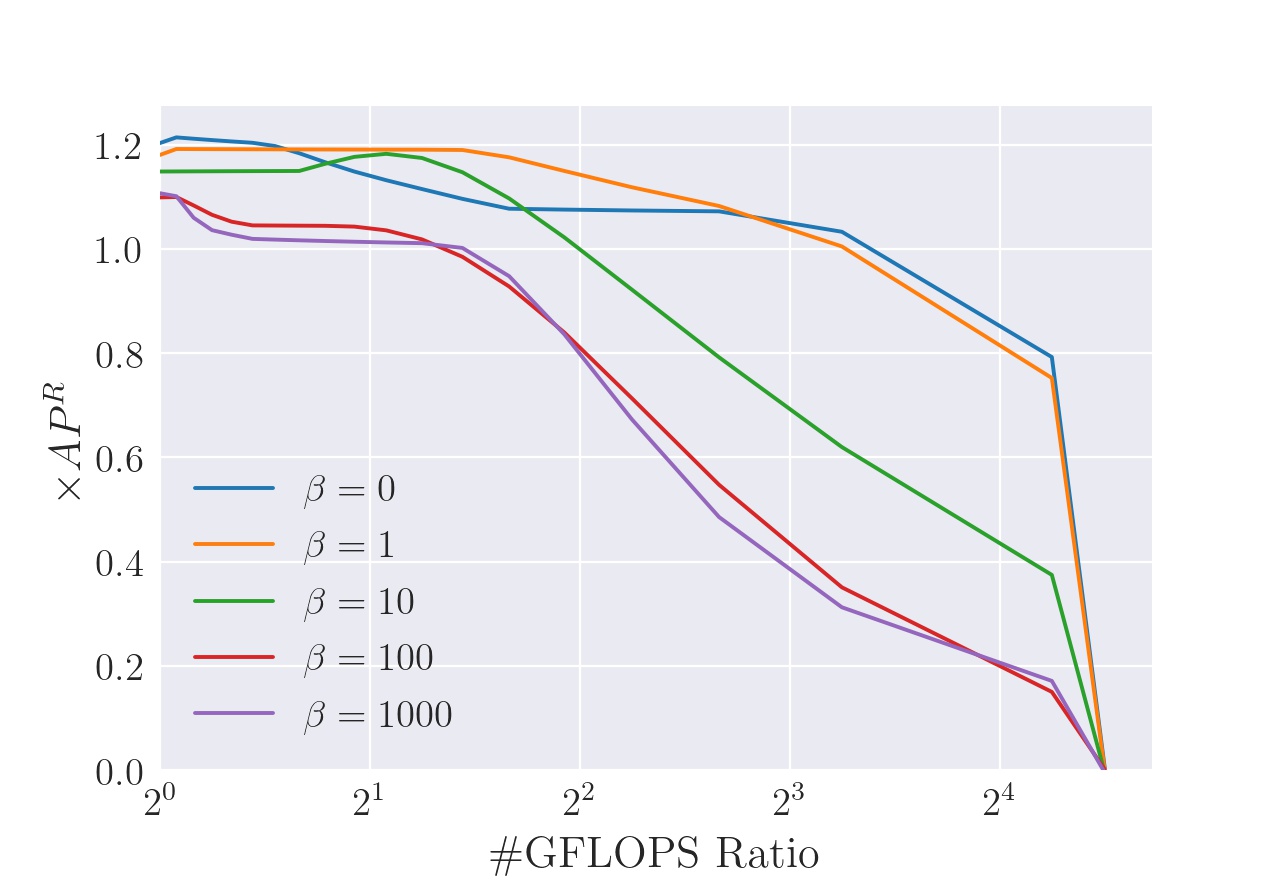}
        \caption{\textbf{$\times AP^R$}. \label{fig:caviar_ap_r_vs_gflops_beat_search}} 
    \end{subfigure}
    \begin{subfigure}[t]{0.49\linewidth}
        \includegraphics[width=1.0\linewidth]{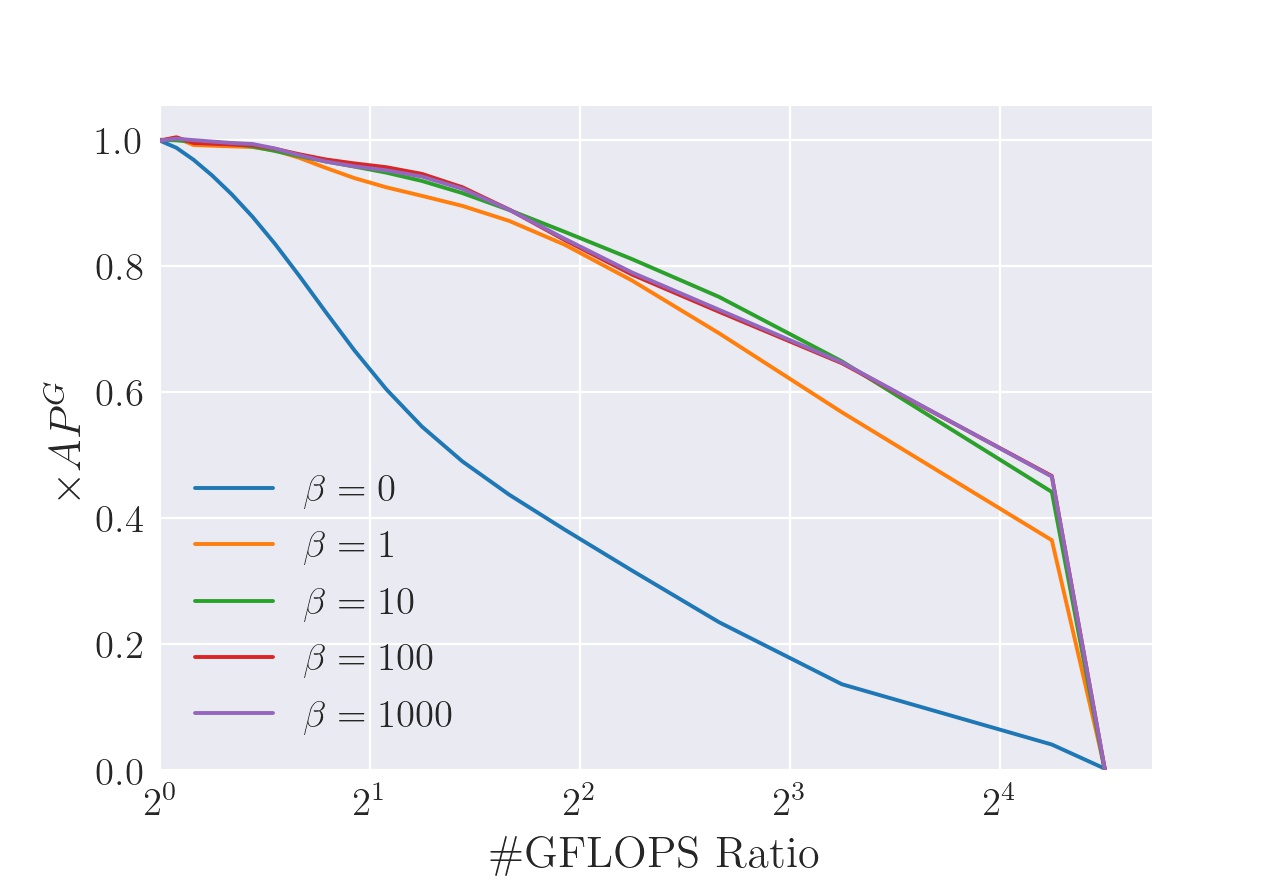}
        \caption{\textbf{$\times AP^G$.} \label{fig:caviar_ap_g_vs_gflops_beat_search}}
    \end{subfigure}
    \vspace{10pt}
    \caption{\textbf{$\beta$ evaluation.} Specialized detector results on the CAVIAR "corridor" camera for various $\beta$ values. Results are obtained by using Eq.~\eqref{eq:loss:final} with our Flops loss (Eq.~\eqref{eq:loss:compression_gflops}).
    }
    \label{fig:example}
\end{figure}

\subsection{Exploration of $\beta$}
We start by analyzing the influence of $\beta$ on the detector performance. From Fig.~\ref{fig:caviar_ap_r_vs_gflops_beat_search} and Fig.~\ref{fig:caviar_ap_g_vs_gflops_beat_search} we see that a value of $\beta=1$ gives good performance both on the generalized and restricted domain. Unsurprisingly, the more the network is pruned, the more its average precision on the general domain drops.

We also tested the influence of $\beta$ on the tracker performance, as can be seen from Fig. ~\ref{fig:beta_tracker}. 
Here $\beta$ weighs the supervised loss on the generalized set as indicated by Eq. \eqref{eq:loss:final}. As can be seen, at $\beta=2$ the best performance is achieved. Since the tracker trains on samples drawn from frames it collects, $\beta$ may serve as a measure to prevent the tracker from possible drift in the following frames. Moreover, the general MDNet model has a 90.9\% precision score and 0.678 success AUC. It is shown that we can compress the tracker specialized model by a rate of $\times$13
with a drop of 1.7\% at the precision score and 2.7\%  at the success score.

\section{Conclusions}
\label{sec:conc}
We propose a novel, self-supervised, approach for specializing and compressing models to restricted domains where compute and storage are limited but so is the environment. 
Our approach provides control over the accuracy obtained in the specialized domain, the amount of generalized information lost as part of the specialization, and the compression the models can achieve, as well as balance compression and runtime. 
To our knowledge, although others have certainly proposed methods providing each of these benefits separately, we are unaware of previous work which aims to optimize over all of them jointly.

\bibliography{references}
\end{document}